 \documentclass[pmlr,twocolumn,10pt]{jmlr} 

\usepackage{float}

\usepackage{booktabs}
\usepackage{caption}
\usepackage{tcolorbox}
\usepackage{enumitem}
\usepackage{graphicx}

\usepackage{siunitx}
\usepackage{dblfloatfix} 
\usepackage{placeins}
\usepackage[switch]{lineno}

\theorembodyfont{\upshape}
\theoremheaderfont{\scshape}
\theorempostheader{:}
\theoremsep{\newline}

\jmlrvolume{297}
\jmlryear{2025}
\jmlrworkshop{Machine Learning for Health (ML4H) 2025} 

\DeclareRobustCommand{\RadGameLogo}{%
  \raisebox{-0.3\height}{\includegraphics[width=0.8cm]{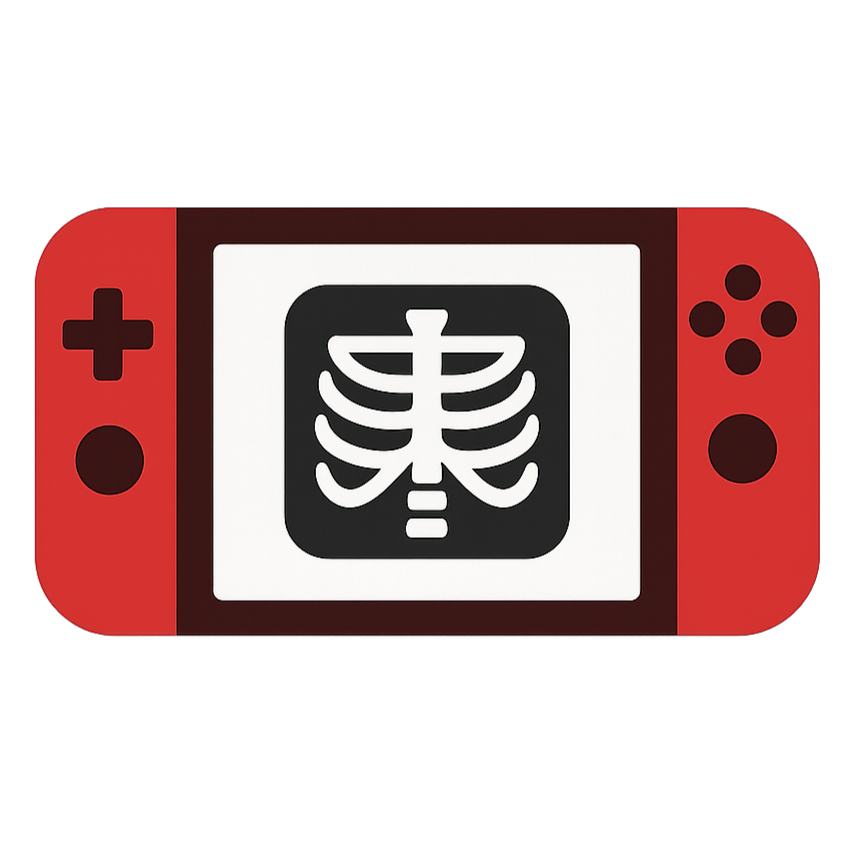}}%
}

\title[RadGame]{%
  \texorpdfstring{\RadGameLogo\ RadGame: An AI-Powered Platform for Radiology Education}%
                  {RadGame: An AI-Powered Platform for Radiology Education}%
}

\newcommand{\addrharvard}{Department of Biomedical Informatics, Harvard Medical School, Boston, MA}
\newcommand{\addrbwh}{Department of Radiation Oncology, Mass General Brigham, Boston, MA}
\newcommand{\addrmaastricht}{Maastricht University, Maastricht, Netherlands}

\newcommand{\addrngha}{Department of Medical Imaging, King Abdulaziz Medical City, Ministry of National Guard, Riyadh, Saudi Arabia}
\newcommand{\addrsnu}{National Strategic Technology Research Institute, Seoul National University Hospital, South Korea}
\newcommand{\addrslu}{Saint Louis University School of Medicine, St. Louis, MO}
\newcommand{\addrksauhs}{College of Medicine, King Saud bin Abdulaziz University for Health Sciences, Riyadh, Saudi Arabia}
\newcommand{\addrkaimrc}{Department of Data Management, King Abdullah International Medical Research Center, Riyadh, Saudi Arabia}
\newcommand{\addrksbh}{Department of Health Informatics, King Saud Bin Abdulaziz University for Health Sciences, Riyadh, Saudi Arabia}
\newcommand{\addrbidmc}{Beth Israel Deaconess Medical Center, Boston, MA}
\newcommand{\addrkfshrc}{Diagnostic Imaging Department, King Faisal Specialist Hospital \& Research Center, Riyadh, Saudi Arabia}
\newcommand{\addmassgen}{Department of Dermatology, Massachusetts General Hospital, Boston, MA}

\author{%
\Name{Mohammed Baharoon}$^{*,1}$, 
\Name{Siavash Raissi}$^{*,1}$, 
\Name{John S. Jun}$^{*,1}$, 
\Name{Thibault Heintz}$^{*,2,3}$, 
\Name{Mahmoud Alabbad}$^{4}$, 
\Name{Ali Alburkani}$^{4}$, 
\Name{Sung Eun Kim}$^{1,5}$, 
\Name{Kent Kleinschmidt}$^{6}$, 
\Name{Abdulrahman O. Alhumaydhi}$^{7}$, 
\Name{Mohannad Mohammed G. Alghamdi}$^{7}$, 
\Name{Jeremy Francis Palacio}$^{6}$, 
\Name{Mohammed Bukhaytan}$^{7}$, 
\Name{Noah Michael Prudlo}$^{6}$, 
\Name{Rithvik Akula}$^{6}$, 
\Name{Brady Chrisler}$^{6}$, 
\Name{Benjamin Galligos}$^{6}$, 
\Name{Mohammed O. Almutairi}$^{7}$, 
\Name{Mazeen Mohammed Alanazi}$^{7}$, 
\Name{Nasser M. Alrashdi}$^{7}$, 
\Name{Joel Jihwan Hwang}$^{6}$, 
\Name{Sri Sai Dinesh Jaliparthi}$^{6}$, 
\Name{Luke David Nelson}$^{6}$, 
\Name{Nathaniel Nguyen}$^{6}$, 
\Name{Sathvik Suryadevara}$^{6}$, 
\Name{Steven Kim}$^{8}$, 
\Name{Mohammed F. Mohammed}$^{9}$, 
\Name{Yevgeniy R. Semenov}$^{10}$, 
\Name{Kun-Hsing Yu}$^{1}$, 
\Name{Abdulrhman Aljouie}$^{11,12}$, 
\Name{Hassan AlOmaish}$^{\dagger,4}$, 
\Name{Adam Rodman}$^{\dagger,13}$, 
\Name{Pranav Rajpurkar}$^{\dagger,1}$ 
\Email{mohammed\_baharoon@hms.harvard.edu} \\
\\
$^1$ \addr \addrharvard \\
$^2$ \addr \addrbwh \\
$^3$ \addr \addrmaastricht \\
$^4$ \addr \addrngha \\
$^5$ \addr \addrsnu \\
$^6$ \addr \addrslu \\
$^7$ \addr \addrksauhs \\
$^8$ Tufts University School of Medicine, Boston, MA \\
$^9$ \addr \addrkfshrc \\
$^{10}$ \addr \addmassgen \\
$^{11}$ \addr \addrkaimrc \\
$^{12}$ \addr \addrksbh \\
$^{13}$ \addr \addrbidmc \\
\AND
$^*$ These authors contributed equally. \\
$^\dagger$ Senior authors.
}

\begin{document}

\maketitle

\begin{abstract}
We introduce \textbf{RadGame}, an AI-powered gamified platform for radiology education that targets two core skills: localizing findings and generating reports. Traditional radiology training is based on passive exposure to cases or active practice with real-time input from supervising radiologists, limiting opportunities for immediate and scalable feedback. RadGame addresses this gap by combining gamification with large-scale public datasets and automated, AI-driven feedback that provides clear, structured guidance to human learners. In \textit{RadGame Localize}, players draw bounding boxes around abnormalities, which are automatically compared to radiologist-drawn annotations from public datasets, and visual explanations are generated by vision-language models for user missed findings. In \textit{RadGame Report}, players compose findings given a chest X-ray, patient age and indication, and receive structured AI feedback based on radiology report generation metrics, highlighting errors and omissions compared to a radiologist's written ground truth report from public datasets, producing a final performance and style score. In a prospective evaluation, participants using RadGame demonstrated a \textit{68\%} improvement in localization accuracy compared to \textit{17\%} with traditional passive methods and a \textit{31\%} improvement in report-writing accuracy compared to \textit{4\%} with traditional methods after seeing the same cases. RadGame highlights the potential of AI-driven gamification to deliver scalable, feedback-rich radiology training and reimagines the application of medical AI resources in education.
\end{abstract}


\paragraph*{Data and Code Availability}

RadGame is built upon publicly available radiology datasets. For \textit{RadGame Localize}, we used the PadChest-GR dataset \citep{de2025padchest}, which contains chest radiographs with radiologist-annotated bounding boxes. For \textit{RadGame Report}, we used the ReXGradient-160K dataset \citep{zhang2025rexgradient}, which provides X-rays paired with radiologist-written reports. Both datasets are freely available to the research community under their respective licenses. 

The RadGame platform code is avaliable at \url{https://github.com/siavashraissi/RadGame}.

\paragraph*{Institutional Review Board (IRB)}

This study was reviewed by the Harvard Faculty of Medicine Institutional Review Board (Protocol \#IRB25-0694) and determined to be exempt under 45 CFR 46.104(d)(2)(3).

\section{Introduction}
\label{sec:intro}

Radiology trainees must acquire two fundamental skills: accurately identifying abnormalities on imaging studies and articulating findings in clear, structured reports. Traditional approaches to radiology education rely on didactic lectures, passive exposure to cases, and supervised readings \citep{griffith2019radiology}. These methods provide limited opportunities for immediate, personalized feedback. Prior work in medical education has shown that active learning improves diagnostic accuracy and knowledge retention, suggesting the need for more interactive and feedback-rich training methods in radiology \citep{freeman2014active}. Moreover, recent work on artificial intelligence (AI) augmented education further highlights its potential to deliver adaptive, real-time feedback and personalized learning experiences across medical training settings \citep{shaw2025artificial, hui2025exploring}.

However, current radiology education platforms fall short in two key ways. First, most lack real-time, structured, and personalized feedback: trainees can review cases or observe ground-truth annotations, but rarely receive automated guidance tailored to their specific errors or skill level \citep{duong2019artificial, griffith2019radiology}. Second, existing platforms often rely on small, highly curated datasets for simplified tasks that do not capture the diversity, complexity, or volume of real-world radiology interpretation \citep{biswas2022current, banerjee2023radhunters, ali2021sonogames}. This limits their ability to provide trainees with the breadth of exposure and adaptive learning experiences needed to prepare for real-world clinical practice.
 

To address these limitations, we report the following contributions:
(1) We develop \textit{RadGame}, an AI-powered gamified platform that teaches the two core radiology tasks—localization and report writing—by repurposing existing large-scale AI radiology datasets and evaluation metrics to provide structured feedback for human learners. (2) We conduct a prospective, multi-institutional user study, showing that using RadGame is associated with improvements in localization and report-writing performance compared to traditional passive learning. (3) We develop \textit{CRIMSON}, an extension of GREEN motivated by RadGame’s role as a human-centered evaluation framework. RadGame revealed GREEN’s limitations in accounting for clinical context (e.g., age, indication) when measuring the clinical significance of errors, leading to CRIMSON as a more context-aware metric.

\begin{figure*}[!h]
  \centering
  \includegraphics[width=.95\textwidth]{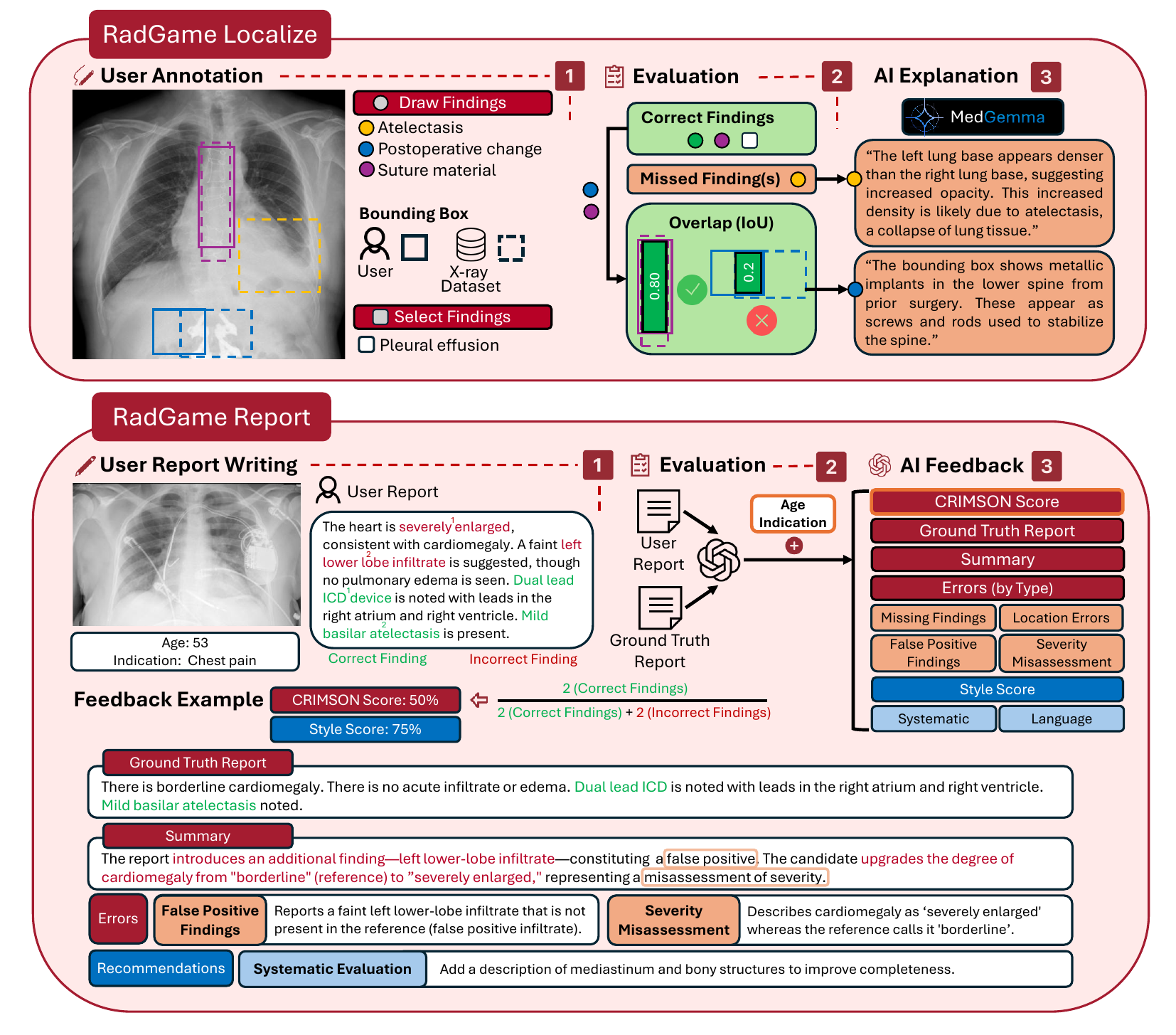}
  \caption{\textbf{Overview of RadGame's User Workflow.} In \textit{Localize}, users identify chest X-ray findings either by drawing bounding boxes for location-dependent abnormalities (Draw findings) or by selecting findings that are consistently associated with a fixed anatomical region or that cannot be localized (Select findings). For all existing findings, ground truth bounding boxes are overlaid, and MedGemma 4B generates explanations for findings that are missed or incorrectly identified. A finding is considered correct only if the IoU is over 0.25. In \textit{Report}, users draft finding reports that are assessed by a GPT-o3 using CRIMSON (see Section \ref{sec:crimson}), producing structured outputs that include the CRIMSON score, ground truth findings, summary, and categorized errors. A Style Score is also produced, which covers the report's completeness across all major chest X-ray regions (lungs, heart, bones, mediastinum) and the use of full sentences and clinical language.}
  \
  \label{fig:main_figure}
\end{figure*}

\section{Related Works}

\begin{figure*}[!h]
  \centering
  \includegraphics[width=0.95\textwidth]{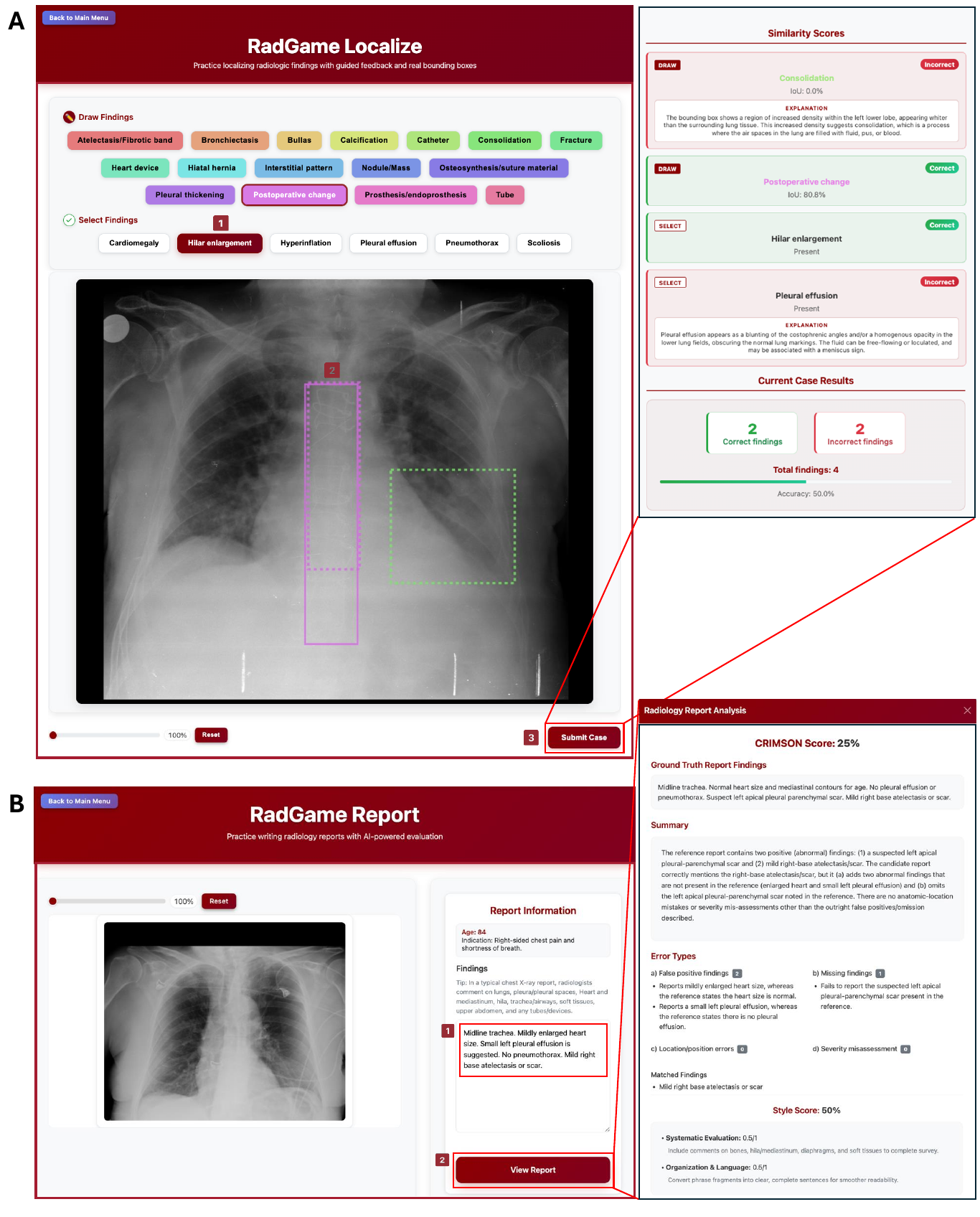}
  \caption{\textbf{RadGame User Interface.} Screenshot of the RadGame platform showing both modules: \textbf{(A)} \textit{Localize}, where users identify findings on chest X-rays either by drawing bounding boxes or selecting predefined options, and \textbf{(B)} \textit{Report}, where users compose findings reports given the image, age, and indication. }
  \label{fig:ui_screenshot}
\end{figure*}

Radiology education has traditionally relied on static textbooks, didactic lectures, and limited opportunities for interactive feedback. In recent years, there has been growing interest in incorporating gamification and personalized training paradigms to better align with the needs of modern trainees and the cognitive demands of radiologic interpretation \citep{duong2019artificial, hui2025exploring, biswas2022current}.

Several efforts have explored gamified systems for radiology learning. Banerjee et al. introduced RADHunters, a first-person game simulating nodule detection tasks in chest CT imaging, showing that gamification can enhance perceptual skill acquisition and learner engagement \citep{banerjee2023radhunters}. Similarly, SonoGames utilized competitive ultrasound-based quizzes in residency training programs, demonstrating improvements in knowledge retention \citep{ali2021sonogames}. \cite{winkel2020gamification} evaluated a gamified e-learning platform for pneumothorax detection, where timed challenges with immediate feedback significantly increased diagnostic confidence and skill retention among radiology residents. Additionally, \cite{mobley2023impact} reviewed the role of the Kaizen platform, an app-based educational tool with gamified multiple-choice formats, instant feedback, and competitive leaderboards to motivate learners. Early evidence suggests Kaizen enhances knowledge retention, peer interaction, and engagement among radiology trainees.

Recent work has also explored integrating artificial intelligence into radiology education \citep{hui2025exploring, wang2024application, cheng2020artificial, saricilar2023pilot}. \cite{biswas2022current} demonstrated the potential of AI-augmented education through a web-based application for chest X-ray nodule detection, showing that AI can deliver real-time, interactive feedback to enhance perceptual training. Similarly, \cite{cheng2020artificial} showed that an AI-assisted education system significantly improved medical students’ diagnostic accuracy for hip fracture detection, particularly among students with lower baseline performance.

Building on these findings, RadGame proposes the first AI-driven educational platform that combines interactive finding localization across 22 finding class types and report writing with automated, personalized feedback at scale.

\begin{figure*}[!t]
  \centering
  \includegraphics[width=\textwidth]{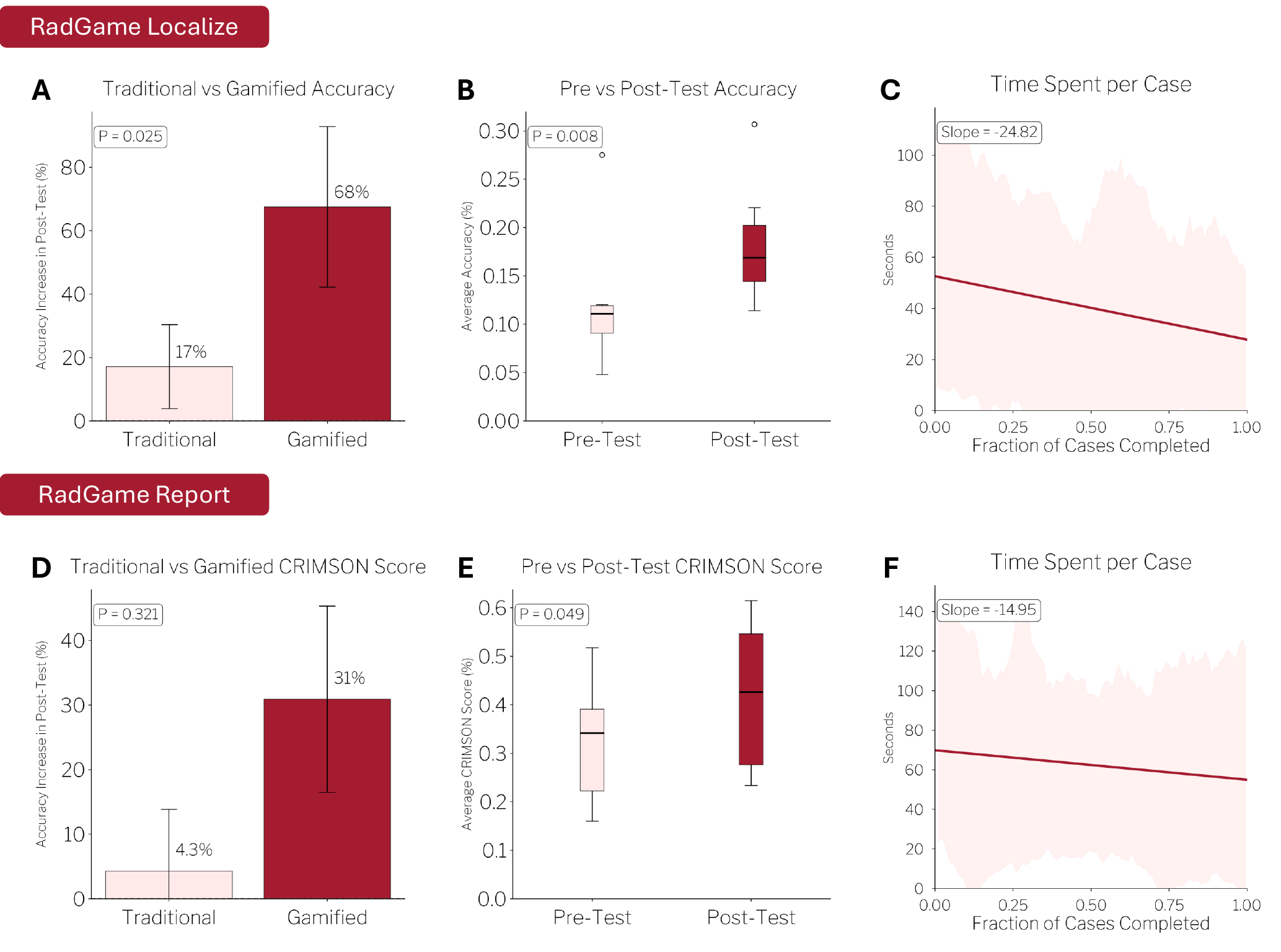}
  \caption{\textbf{Performance and efficiency improvements with RadGame across both modules.} The top row shows results for RadGame Localize: \textbf{(A)} comparison of accuracy improvements between Gamified and Traditional groups, \textbf{(B)} pre-test vs. post-test accuracy changes, and \textbf{(C)} reduction in time spent per case over training. The bottom row shows corresponding results for RadGame Report: \textbf{(D)} accuracy improvements in Gamified vs. Traditional groups, \textbf{(E)} pre-test vs. post-test CRIMSON score changes, and \textbf{(F)} reduction in time spent per case. Statistical significance for Gamified vs. Traditional comparisons was assessed using a two-tailed Mann–Whitney U test, while pre- vs. post-test comparisons were assessed using a one-tailed Wilcoxon signed-rank test. Error bars represent the Standard Error of the Mean.}
  \
  \label{fig:results}
\end{figure*}

\section{RadGame}\label{sec:radgame}

\textit{RadGame} is an AI-powered gamified platform that integrates two core radiology tasks—report generation and finding localization—into interactive learning modules. The platform leverages large-scale public datasets and AI-based feedback mechanisms to provide trainees with structured, personalized evaluation at scale. RadGame consists of two modules: \textit{RadGame Localize} and \textit{RadGame Report}. Figure \ref{fig:main_figure} shows the overall workflow of the two modules.

\subsection{RadGame Localize}
In \textit{RadGame Localize}, participants are shown single frontal chest X-rays from the PadChest-GR dataset \citep{de2025padchest} and are prompted to identify different radiological findings. Since finding labels in PadChest-GR come directly from reports, they were originally very specific, with classes like ``lobar atelectasis" and ``segmental atelectasis," which led to some labels having a very small occurrence count. These findings were combined into a single class (for example, ``atelectasis"). They are then divided into two categories: \textit{Draw Findings} and \textit{Select Findings}. Supplementary Figure \ref{tab:findings} shows all the classes.

\textit{Draw Findings} are abnormalities that require the trainee not only to identify their presence but also to localize them by drawing a bounding box on the image. These findings are typically focal and location-dependent, meaning they can appear in different regions of the chest (``Nodule/Mass", ``Fracture", ``Calcification", etc.). A prediction is considered correct if the intersection-over-union (IoU) between the trainee’s bounding box and the radiologist’s ground-truth annotation exceeds 0.25, a threshold determined in consultation with radiologists. \textit{Select Findings}, in contrast, are findings where trainees only need to indicate their presence from a checklist without drawing a box. These are typically diffuse abnormalities or findings that are consistently associated with a fixed anatomical location (``Cardiomegaly", ``Scoliosis", etc.).  

When a trainee misses a finding, visual feedback is provided through MedGemma 4B \citep{sellergren2025medgemma}. For Draw Findings, the system overlays the ground-truth bounding box on the image and generates a two-sentence explanation describing how the missed abnormality can be visually recognized. For Select Findings, MedGemma provides a general description of the typical radiographic appearance of the missed finding. An example is shown in Figure \ref{fig:ui_screenshot}A. In Appendix \ref{sec:medgemma_explinations_val} we provide a validation of MedGemma's explanations. 

\subsection{RadGame Report}

In \textit{RadGame Report}, trainees are presented with all images from a chest X-ray study and prompted to write a radiology report. The ground-truth reference is derived from the \textit{Findings} section of the ReXGradient-160K dataset \citep{zhang2025rexgradient}. Studies that have priors or comparisons are excluded. Submitted reports are automatically evaluated using the CRIMSON metric (Section~\ref{sec:crimson}), a report generation metric adapted from GREEN \citep{ostmeier2024green}. Unlike GREEN, CRIMSON ignores normal findings that would otherwise inflate scores and incorporates patient age and clinical indication to weigh the clinical significance of errors. Formally, CRIMSON is defined by adapting the GREEN score formula:
\[
\text{Score} =
\frac{\# \,\text{matched findings}}
     {\# \,\text{matched findings} \;+\;
      \sum_{i=(a)}^{(d)} \# \,\text{error}_{\text{sig}, i}}
\]
where \# matched findings is the number of findings that appear in both the candidate and the reference report and errors \textit{a} through \textit{d} corresponding to the four different error types (see Figure \ref{fig:main_figure}). Errors \textit{e} (mentioning a comparison that isn't in the reference) and \textit{f} (omitting a comparison detailing a change from a prior study) from GREEN are dropped since they are concerned with priors. 

We use GPT-o3 as our LLM model to generate the report evaluation. The evaluation output includes a single CRIMSON score between 0 and 100\%, the ground truth report for comparison, a structured summary of the trainee’s submission and errors, and CRIMSON error categories that cover four error categories: false positives, missing findings, location or position errors, and severity misclassification. Figure \ref{fig:ui_screenshot}B shows an example of this.

The evaluation output also includes a ``Style Score", scaled from 0 (a poorly styled report) to 100\% (a professionally styled report), encouraging trainees to use best practices in radiology reporting. The style score considers whether the report covers all major chest X-ray regions (lungs, heart, bones, mediastinum), as well as the organization of the report including use of complete sentences and clinical language (see Supplementary Figure \ref{fig:stylescore-prompt} for the prompt).

\subsection{Customizability}
RadGame is designed with flexibility in mind, allowing adaptation across multiple dimensions. The platform can be extended to incorporate different datasets, additional finding categories, and alternative feedback strategies. For the localization task in particular, RadGame supports customizing the set of \textit{Draw} and \textit{Select} findings to match the characteristics of the dataset or the learning objectives. For example, the platform can be easily extended to support a version for localizing nodules only.

To further support this extensibility, we developed a specialized version of RadGame focused on interstitial lung patterns, which are subtle and diagnostically challenging findings. For this module, we created the first bounding-box annotated dataset based on ReXGradient-160k \citep{zhang2025rexgradient} specifically aimed at differentiating between distinct interstitial patterns, including Kerley lines, miliary, and reticulonodular opacities. The distribution of findings for this dataset is shown in Supplementary Table \ref{tab:interstitial_patterns}. The dataset will be made publicly available.

\section{User Study}\label{sec:userstudy}

\subsection{Study Design}
\textbf{Cohort.} Eighteen medical students were selected to participate in the user study (see Supplementary Table \ref{tab:demographics} for demographics). Participants come from Saint Louis University (Saint Louis, MO), Tufts University School of Medicine (Boston, MA), and King Saud bin Abdulaziz University for Health Sciences (Riyadh, Saudi Arabia), comprising a multinational, multi-institutional cohort. Students were selected based on their interest in radiology. At the time of participation, nine students were in the pre-clinical stage (M1, M2) of medical training and nine were in the clinical stage (M3, M4). When asked about prior radiology experience across classes and clinical rotations, seven students (38.8\%) reported no experience, five (27.8\%) had only taken radiology classes, two (11.1\%) had only performed radiology rotations, and four (22.2\%) had experience with both.

\textbf{Group Assignment.} Participants first completed the \textit{RadGame Localize} module, followed by \textit{RadGame Report} module. They were randomly assigned to either a \textit{Gamified} or \textit{Traditional} feedback group for each module, with a crossover design such that participants assigned to the Gamified group for Localize were assigned to the Traditional group for Report, and vice versa. This resulted in 8 and 10 participants completing Localize and Report, respectively, in the Gamified group.

Participants assigned to the Gamified group received AI-generated explanations and context-aware error feedback, while those in the Traditional modality learned by observing cases and ground truths without interactive feedback (see Supplementary Figure \ref{fig:screen_shots_passive}). For \textit{RadGame Localize}, the Gamified group drew bounding boxes with AI-generated explanations for user missed findings (see Figure \ref{fig:ui_screenshot}A), whereas the Traditional group viewed only the case with ground-truth boxes overlaid. For \textit{RadGame Report}, the Gamified group wrote reports with full automated and personalized feedback, while the Traditional group viewed only the chest X-ray and the associated ground-truth report (see Figure \ref{fig:ui_screenshot}B). Both groups saw the same cases in the same order.

\textbf{Study phases.} The study was conducted in three phases: a baseline pre-test, a learning phase with the assigned group, and a post-test with identical cases to the pre-test for final evaluation. Participants were asked to complete the three components within a seven-day period and did not receive their test scores during the pre- and post-tests.

In RadGame Localize, the pre-test and post-test consisted of completing 25 RadGame Localize cases completed within a 45-minute timeframe. A senior radiologist selected the test cases from PadChest-GR to evenly distribute case difficulties \citep{de2025padchest}. To ensure that the ground truth annotations of the test cases were as accurate and consistent as possible, two senior radiologists were assigned to re-evaluate the findings and their respective annotations\textemdash revising bounding boxes while adding missed conditions. Upon completion of the pre-test, participants were assigned to complete 375 cases of RadGame Localize in their assigned group. Once participants completed all cases, they were asked to take the post-test and move to RadGame Report.

In RadGame Report, the pre-test and post-test consisted of completing 10 RadGame Report cases within a 45-minute timeframe. A senior radiologist selected the test cases from ReXGradient-160K to establish an even distribution of case difficulties \citep{zhang2025rexgradient}. Ground-truth and student reports for both tests were reviewed by radiologists to ensure that final scores reliably measured students' abilities to write reports, rather than optimize CRIMSON scores. The review procedure is described further in Appendix \ref{sec:reportvalidation}. Upon completion of the pre-test, participants were assigned to complete 150 cases of RadGame Report in their assigned group. Once participants completed all cases, they were asked to take the post-test to complete the study.

\subsection{Results}

We evaluated the impact of RadGame on diagnostic accuracy and efficiency across both the \textit{Localize} and \textit{Report} modules (Figure~\ref{fig:results}). Across both modules, participants assigned to the \textit{Gamified} group demonstrated larger performance gains than those in the \textit{Traditional} modality.

For RadGame Localize, participants in the Gamified group showed a 68\% improvement in post-test accuracy relative to baseline, compared to only 17\% in the Traditional group ($p<0.05$, two-tailed Mann--Whitney U). Figure \ref{fig:results}B shows a comparison between pre- vs. post-test gains for the Gamified group ($p<0.05$, one-tailed Wilcoxon signed-rank). Specifically, the pre-test showed a median accuracy of 0.111 (range: 0.048–0.275, 95\% CI: 0.076–0.120), while the post-test showed a median of 0.169 (range: 0.114–0.307, 95\% CI: 0.138–0.221), corresponding to a Cliff’s Delta of 0.673 compared to the pre-test. These results demonstrate that interactive bounding box annotation with AI-generated feedback was associated with improved localization skills compared to traditional passive review of ground-truth annotations. In Supplementary Figure \ref{fig:localize_thresholds}, we also show localize accuracies across different IoU thresholds, and Supplementary Figure \ref{fig:localize_label_accuracy} shows accuracy differences in the pre- and post-test across finding labels.

For RadGame Report, the Gamified group showed a 31\% improvement in CRIMSON scores from pre- to post-test, compared to 4.3\% in the Traditional group. While this between-group difference did not reach statistical significance, this is likely due to the small sample size (n = 18) despite the large observed difference. Figure \ref{fig:results}E revealed significant pre- vs. post-test improvements in the Gamified group ($p<0.05$). Specifically, the pre-test showed a median CRIMSON score of 0.342 (range: 0.160–0.518, 95\% CI: 0.222–0.412), while the post-test showed a median of 0.426 (range: 0.233–0.615, 95\% CI: 0.268–0.547), with a Cliff’s Delta of 0.383 compared to the pre-test. These findings demonstrate that receiving personalized feedback from RadGame may be associated with improved report writing skills.

Finally, across both modules, participants in the Gamified group demonstrated progressive reductions in time spent per case as training advanced, indicating improved diagnostic efficiency alongside accuracy gains (Figure~\ref{fig:results}C and F). At the end of the training, participants were around 25 seconds faster per case at RadGame Localize and 15 seconds faster per case at RadGame Report. A plot comparing time spent for Gamified and Traditional groups is shown in Supplementary Figure \ref{fig:time_distributions}.

\section{RadGame as a Humanistic Evaluation of AI Material}\label{sec:crimson}

Beyond its role as an educational tool, \textit{RadGame} can act as a human-in-the-loop evaluation harness for AI  radiology material—analogous to Chatbot Arena for large language models \citep{chiang2024chatbot}. In our setting, trainees interact directly with material used to train and evaluate AI models and provide implicit and explicit judgments through gameplay, which serves as a humanistic complement to correlation-based benchmarks.

This was exemplified during our user study, where user feedback post-study revealed two key shortcomings of the GREEN metric, which was developed to evaluate AI report generation capabilities \citep{ostmeier2024green}. First, GREEN rewarded the reporting of normal findings, inflating scores even when significant abnormalities were missed. However, the inclusion of specific normal findings in a report is often determined by the individual radiologist’s style rather than by clinical necessity, making such matches a poor basis for evaluation of clinical accuracy. Instead, we believe that it makes more sense to keep a separate score for evaluating report style. This led to us proposing ``Style Score," a separate scoring system that evaluates a report along two pillars: (1) Systematic Evaluation and (2) Organization and Language. Systematic evaluation verifies if the report systematically evaluates all important regions in a chest x-ray, including the lungs, heart, bones, and mediastinum. Organization and Language evaluates the report for the use of clinical language, writing in full sentences, and having a generally organized report. The prompt can be found in Supplementary Figure \ref{fig:stylescore-prompt}. Second, GREEN fails to account for clinical context, penalizing omissions of irrelevant findings (e.g., age-related degenerative changes) while giving equal weight to errors that were clinically consequential. 

Guided by these insights, we introduced CRIMSON, a novel metric that (1) ignores matches on normal findings and (2) incorporates age and indication to assess the clinical significance of errors. Figure~\ref{fig:crimson} illustrates these improvements. The prompt for CRIMSON can be found in Supplementary Figure \ref{fig:CRIMSON-prompt}. In this way, RadGame not only trains human learners but also functions as a testbed for refining AI evaluation frameworks, bridging the gap between quantitative evaluations and human judgment.

\begin{figure}[h]
  \centering
  \includegraphics[width=\columnwidth]{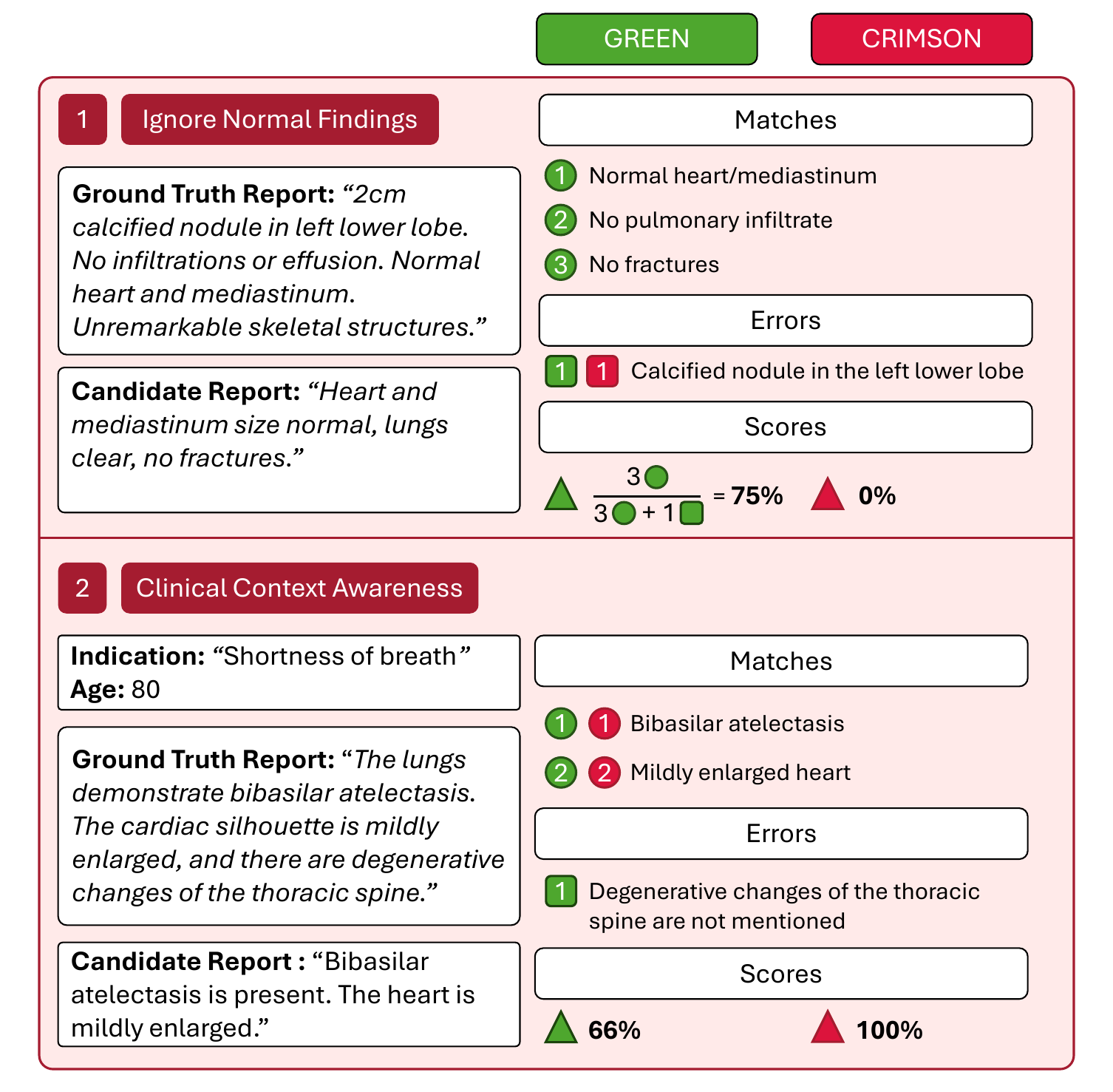}
  \caption{\textbf{Comparison of GREEN and CRIMSON scoring.} 
  (1) \textit{Ignore Normal Findings:} GREEN rewards normal findings (e.g., normal heart/mediastinum, no infiltrate, no fractures), inflating the score despite missing the clinically important calcified nodule. CRIMSON excludes such credit, yielding 0\%. 
  (2) \textit{Clinical Context Awareness:} For an 80-year-old with shortness of breath, GREEN penalizes omission of degenerative spine changes, whereas CRIMSON deems them insignificant, giving full credit for bibasilar atelectasis and mild cardiomegaly.}
  \label{fig:crimson}
\end{figure}

\section{Discussion}

This work introduces RadGame, an AI-powered, gamified platform for radiology education that integrates real-time, structured feedback into both localization and report-writing tasks. Our findings demonstrate that gamified, feedback-rich training is associated with significantly larger gains in diagnostic accuracy and efficiency compared to traditional passive learning methods. These results align with a growing body of literature showing that active learning paradigms and immediate feedback may help with diagnostic reasoning and skill retention across medical education domains \citep{freeman2014active, duong2019artificial, banerjee2023radhunters}. By repurposing large-scale, publicly available radiology datasets and AI evaluation resources into human-centered training experiences, RadGame builds upon earlier efforts in gamification and AI-augmented education while uniquely scaling structured feedback to diverse tasks.

Future efforts will expand RadGame to incorporate additional imaging modalities and tasks beyond chest X-rays. In particular, grounding datasets such as ReXGroundingCT \citep{baharoon2025rexgroundingct} offer an opportunity to develop RadGame modules for volumetric CT imaging. Such expansions could support training in cross-sectional imaging and enable assessment of three-dimensional spatial reasoning, which is inherently more difficult. Another direction involves making the platform more interactive through back-and-forth dialogue between the trainee and the AI system. Rather than providing static feedback after a single submission, future iterations could allow learners to ask clarifying questions, receive hints in real time, and iteratively refine their reports or localizations based on the AI’s guidance. Such a conversational framework would transform RadGame into a more adaptive, tutor-like experience, fostering deeper reasoning and personalized learning trajectories.

This study has several limitations. First, the sample size was relatively small (n = 18), which may limit statistical power for some comparisons. However, we partly mitigate this by including a multi-institutional cohort with participants at different stages of medical training and diverse prior experiences in radiology, improving the representativeness of our findings. Second, we observed that participants in the Traditional group progressed through the cases more quickly than those in the Gamified group. This difference reflects the passive nature of the Traditional modality, where participants simply observed the cases and ground-truth annotations without the additional interactive steps required in the Gamified setting.

\acks{This project started as a group project from the BMI 702: Foundations of Biomedical Informatics II course at Harvard University. We thank Dr. Marinka Zitnik for her guidance and support throughout the project and course.}

\newpage

\bibliography{jmlr-sample}

\begin{thebibliography}{19}
\providecommand{\natexlab}[1]{#1}
\providecommand{\url}[1]{\texttt{#1}}
\expandafter\ifx\csname urlstyle\endcsname\relax
  \providecommand{\doi}[1]{doi: #1}\else
  \providecommand{\doi}{doi: \begingroup \urlstyle{rm}\Url}\fi

\bibitem[Ali et~al.(2021)Ali, Nadeem, Khalid, Anwar, Nabie, and Docherty]{ali2021sonogames}
Maria~Fatima Ali, Naila Nadeem, Farah Khalid, Naveed~Muhammad Anwar, Ghulam Nabie, and Charles Docherty.
\newblock Sonogames: sounds of the right kind introducing gamification into radiology training.
\newblock \emph{BMC Research Notes}, 14\penalty0 (1):\penalty0 341, 2021.

\bibitem[Baharoon et~al.(2025)Baharoon, Luo, Moritz, Kumar, Kim, Zhang, Zhu, Alabbad, Alhazmi, Mistry, et~al.]{baharoon2025rexgroundingct}
Mohammed Baharoon, Luyang Luo, Michael Moritz, Abhinav Kumar, Sung~Eun Kim, Xiaoman Zhang, Miao Zhu, Mahmoud~Hussain Alabbad, Maha~Sbayel Alhazmi, Neel~P Mistry, et~al.
\newblock Rexgroundingct: A 3d chest ct dataset for segmentation of findings from free-text reports.
\newblock \emph{arXiv preprint arXiv:2507.22030}, 2025.

\bibitem[Banerjee et~al.(2023)Banerjee, Agarwal, and Auffermann]{banerjee2023radhunters}
Soham Banerjee, Rishabh Agarwal, and William~F Auffermann.
\newblock Radhunters: gamification in radiology perceptual education.
\newblock \emph{Journal of Medical Imaging}, 10\penalty0 (S1):\penalty0 S11905--S11905, 2023.

\bibitem[Biswas et~al.(2022)Biswas, Biswas, Awal, and Goyal]{biswas2022current}
Som~Subhro Biswas, Srirupa Biswas, Sandeep~Singh Awal, and Hitesh Goyal.
\newblock Current status of radiology education online: a comprehensive update.
\newblock \emph{SN comprehensive clinical medicine}, 4\penalty0 (1):\penalty0 182, 2022.

\bibitem[Cheng et~al.(2020)Cheng, Chen, Fu, Chaou, Wu, Hsu, Chang, Chung, Hsieh, Hsieh, et~al.]{cheng2020artificial}
Chi-Tung Cheng, Chih-Chi Chen, Chih-Yuan Fu, Chung-Hsien Chaou, Yu-Tung Wu, Chih-Po Hsu, Chih-Chen Chang, I-Fang Chung, Chi-Hsun Hsieh, Ming-Ju Hsieh, et~al.
\newblock Artificial intelligence-based education assists medical students’ interpretation of hip fracture.
\newblock \emph{Insights into Imaging}, 11\penalty0 (1):\penalty0 119, 2020.

\bibitem[Chiang et~al.(2024)Chiang, Zheng, Sheng, Angelopoulos, Li, Li, Zhu, Zhang, Jordan, Gonzalez, et~al.]{chiang2024chatbot}
Wei-Lin Chiang, Lianmin Zheng, Ying Sheng, Anastasios~N Angelopoulos, Tianle Li, Dacheng Li, Banghua Zhu, Hao Zhang, Michael~I Jordan, Joseph~E Gonzalez, et~al.
\newblock Chatbot arena: an open platform for evaluating llms by human preference.
\newblock In \emph{Proceedings of the 41st International Conference on Machine Learning}, pages 8359--8388, 2024.

\bibitem[de~Castro et~al.(2025)de~Castro, Bustos, Bannur, Hyland, Bouzid, Wetscherek, S{\'a}nchez-Valverde, Jaques-P{\'e}rez, P{\'e}rez-Rodr{\'\i}guez, Takeda, et~al.]{de2025padchest}
Daniel~Coelho de~Castro, Aurelia Bustos, Shruthi Bannur, Stephanie~L Hyland, Kenza Bouzid, Maria~Teodora Wetscherek, Maria~Dolores S{\'a}nchez-Valverde, Lara Jaques-P{\'e}rez, Lourdes P{\'e}rez-Rodr{\'\i}guez, Kenji Takeda, et~al.
\newblock Padchest-gr: A bilingual chest x-ray dataset for grounded radiology report generation.
\newblock \emph{NEJM AI}, 2\penalty0 (7):\penalty0 AIdbp2401120, 2025.

\bibitem[Duong et~al.(2019)Duong, Rauschecker, Rudie, Chen, Cook, Bryan, and Mohan]{duong2019artificial}
Michael~Tran Duong, Andreas~M Rauschecker, Jeffrey~D Rudie, Po-Hao Chen, Tessa~S Cook, R~Nick Bryan, and Suyash Mohan.
\newblock Artificial intelligence for precision education in radiology.
\newblock \emph{The British journal of radiology}, 92\penalty0 (1103):\penalty0 20190389, 2019.

\bibitem[Freeman et~al.(2014)Freeman, Eddy, McDonough, Smith, Okoroafor, Jordt, and Wenderoth]{freeman2014active}
Scott Freeman, Sarah~L Eddy, Miles McDonough, Michelle~K Smith, Nnadozie Okoroafor, Hannah Jordt, and Mary~Pat Wenderoth.
\newblock Active learning increases student performance in science, engineering, and mathematics.
\newblock \emph{Proceedings of the national academy of sciences}, 111\penalty0 (23):\penalty0 8410--8415, 2014.

\bibitem[Griffith et~al.(2019)Griffith, Kadom, and Straus]{griffith2019radiology}
Brent Griffith, Nadja Kadom, and Christopher~M Straus.
\newblock Radiology education in the 21st century: threats and opportunities.
\newblock \emph{Journal of the American College of Radiology}, 16\penalty0 (10):\penalty0 1482--1487, 2019.

\bibitem[Hui et~al.(2025)Hui, Sacoransky, Chung, and Kwan]{hui2025exploring}
Muying~Lucy Hui, Ethan Sacoransky, Andrew Chung, and Benjamin~YM Kwan.
\newblock Exploring the integration of artificial intelligence in radiology education: A scoping review.
\newblock \emph{Current Problems in Diagnostic Radiology}, 54\penalty0 (3):\penalty0 332--338, 2025.

\bibitem[Mobley et~al.(2023)Mobley, Chandora, and Woodard]{mobley2023impact}
Alisa Mobley, Agni Chandora, and Stefanie Woodard.
\newblock The impact of gamification and potential of kaizen in radiology education.
\newblock \emph{Clinical Imaging}, 103:\penalty0 109990, 2023.

\bibitem[Ostmeier et~al.(2024)Ostmeier, Xu, Chen, Varma, Blankemeier, Bluethgen, Md, Moseley, Langlotz, Chaudhari, et~al.]{ostmeier2024green}
Sophie Ostmeier, Justin Xu, Zhihong Chen, Maya Varma, Louis Blankemeier, Christian Bluethgen, Arne Md, Michael Moseley, Curtis Langlotz, Akshay Chaudhari, et~al.
\newblock Green: Generative radiology report evaluation and error notation.
\newblock In \emph{Findings of the Association for Computational Linguistics: EMNLP 2024}, pages 374--390, 2024.

\bibitem[Saricilar et~al.(2023)Saricilar, Burgess, and Freeman]{saricilar2023pilot}
Erin~C Saricilar, Annette Burgess, and Anthony Freeman.
\newblock A pilot study of the use of artificial intelligence with high-fidelity simulations in assessing endovascular procedural competence independent of a human examiner.
\newblock \emph{ANZ Journal of Surgery}, 93\penalty0 (6):\penalty0 1525--1531, 2023.

\bibitem[Sellergren et~al.(2025)Sellergren, Kazemzadeh, Jaroensri, Kiraly, Traverse, Kohlberger, Xu, Jamil, Hughes, Lau, et~al.]{sellergren2025medgemma}
Andrew Sellergren, Sahar Kazemzadeh, Tiam Jaroensri, Atilla Kiraly, Madeleine Traverse, Timo Kohlberger, Shawn Xu, Fayaz Jamil, C{\'\i}an Hughes, Charles Lau, et~al.
\newblock Medgemma technical report.
\newblock \emph{arXiv preprint arXiv:2507.05201}, 2025.

\bibitem[Shaw et~al.(2025)Shaw, Henning, and Webster]{shaw2025artificial}
Kody Shaw, Marcus~A Henning, and Craig~S Webster.
\newblock Artificial intelligence in medical education: a scoping review of the evidence for efficacy and future directions.
\newblock \emph{Medical Science Educator}, pages 1--14, 2025.

\bibitem[Wang et~al.(2024)Wang, Huai, Ma, Jin, Wang, Chen, Sang, and Liu]{wang2024application}
DongXu Wang, BingCheng Huai, Xing Ma, BaiMing Jin, YuGuang Wang, MengYu Chen, JunZhi Sang, and RuiNan Liu.
\newblock Application of artificial intelligence-assisted image diagnosis software based on volume data reconstruction technique in medical imaging practice teaching.
\newblock \emph{BMC Medical Education}, 24\penalty0 (1):\penalty0 405, 2024.

\bibitem[Winkel et~al.(2020)Winkel, Brantner, Lutz, Korkut, Linxen, and Heye]{winkel2020gamification}
David~J Winkel, Philipp Brantner, Jonas Lutz, Safak Korkut, Sebastian Linxen, and Tobias~J Heye.
\newblock Gamification of electronic learning in radiology education to improve diagnostic confidence and reduce error rates.
\newblock \emph{American Journal of Roentgenology}, 214\penalty0 (3):\penalty0 618--623, 2020.

\bibitem[Zhang et~al.(2025)Zhang, Acosta, Miller, Huang, and Rajpurkar]{zhang2025rexgradient}
Xiaoman Zhang, Juli{\'a}n~N Acosta, Josh Miller, Ouwen Huang, and Pranav Rajpurkar.
\newblock Rexgradient-160k: A large-scale publicly available dataset of chest radiographs with free-text reports.
\newblock \emph{arXiv preprint arXiv:2505.00228}, 2025.

\end{thebibliography}

\appendix

\clearpage

\section{MedGemma 4B Explanations Validation}\label{sec:medgemma_explinations_val}

To validate MedGemma's feedback, we sampled 100 random RadGame Localize findings. Two board-certified radiologists reviewed each case and scored MedGemma's descriptions on a binary scale for whether it (1) discussed the correct anatomical location, (2) accurately described the image's visual features, and (3) correctly identified the specific subtype of the finding (e.g., correctly labeling ``Pacemaker" when the class is ``Heart Device"). MedGemma addressed the correct anatomical location in 88\% of findings, correctly identified visual features in 96\%, and specified the correct condition present in 90\%. We also show performance for this task using two other models, Qwen3VL and Llama3.2, which is shown in Supplementary Table \ref{tab:model_accuracies}. Supplementary Figure \ref{fig:med_gemma_explinations} shows three examples of such explanations while Supplementary Figure \ref{fig:med_gemma_failure_explinations} shows three examples of failure cases. Supplementary Figure \ref{fig:explinations_accuracy} shows the accuracy of these explanations across finding categories. 


\section{RadGame Report Test Review}\label{sec:reportvalidation}
To ensure that reports in both tests were accurate and did not reference prior studies that would otherwise be unavailable to the participants, two senior board-certified radiologists reviewed and revised all 10 ground-truth reports for comparison. Then, all pre-test and post-test reports written by participants were scored by CRIMSON, followed by manual review by two board-certified radiologists to generate a final score. Radiologists concluded that 3.93\% of the total errors were clinically insignificant or could not be determined from the image and were removed, 3.26\% reflected correct findings that were penalized by CRIMSON (e.g., labeling an interstitial pattern instead of a Kerley B line), while 0.38\% were errors incorrectly deemed insignificant by the model.

\setcounter{figure}{0}

\begin{table*}[t]
\centering
\captionsetup{name=Supplementary Table}
\caption{\textbf{Study Cohort Demographics (Counts and Percentages)}}
\resizebox{\textwidth}{!}{%
\begin{tabular}{l l r r}
\toprule
Question & Response & $n$ & \% \\
\midrule
What is the name of your university/primary affiliation? & St. Louis University & 11 & 61.1 \\
 & King Saud bin Abdulaziz University for Health Sciences & 6 & 33.3 \\
 & Tufts University School of Medicine & 1 & 5.6 \\
\midrule
What stage of medical training are you currently in? & Pre-Clinical & 9 & 50 \\
 & Clinical & 9 & 50 \\
\midrule
Have you done a radiology rotation/classes? & Classes & 5 & 27.8 \\
 & Rotation & 2 & 11.1 \\
 & Both & 4 & 22.2 \\
 & None & 7 & 38.8 \\
\midrule
Determining comfort in radiology - How comfortable & 1 & 0 & 0 \\ are you with interpreting X-ray images on a scale & 2 & 11 & 61.1 \\of 1-5? (1 being the least comfortable) & 3 & 4 & 22.2 \\
 & 4 & 3 & 16.7 \\
 & 5 & 0 & 0 \\
\midrule
Determining comfort in radiology - How comfortable & 1 & 2 & 11.1 \\ are you with PA/AP chest X-rays specifically, on  & 2 & 4 & 22.2 \\a scale of 1-5? (1 being the least comfortable) & 3 & 7 & 38.9 \\
 & 4 & 3 & 16.7 \\
 & 5 & 0 & 0 \\
\bottomrule
\end{tabular}%
}
\label{tab:demographics}
\end{table*}
\newpage

\begin{figure*}[htbp]
  \centering
  \includegraphics[width=0.95\textwidth]{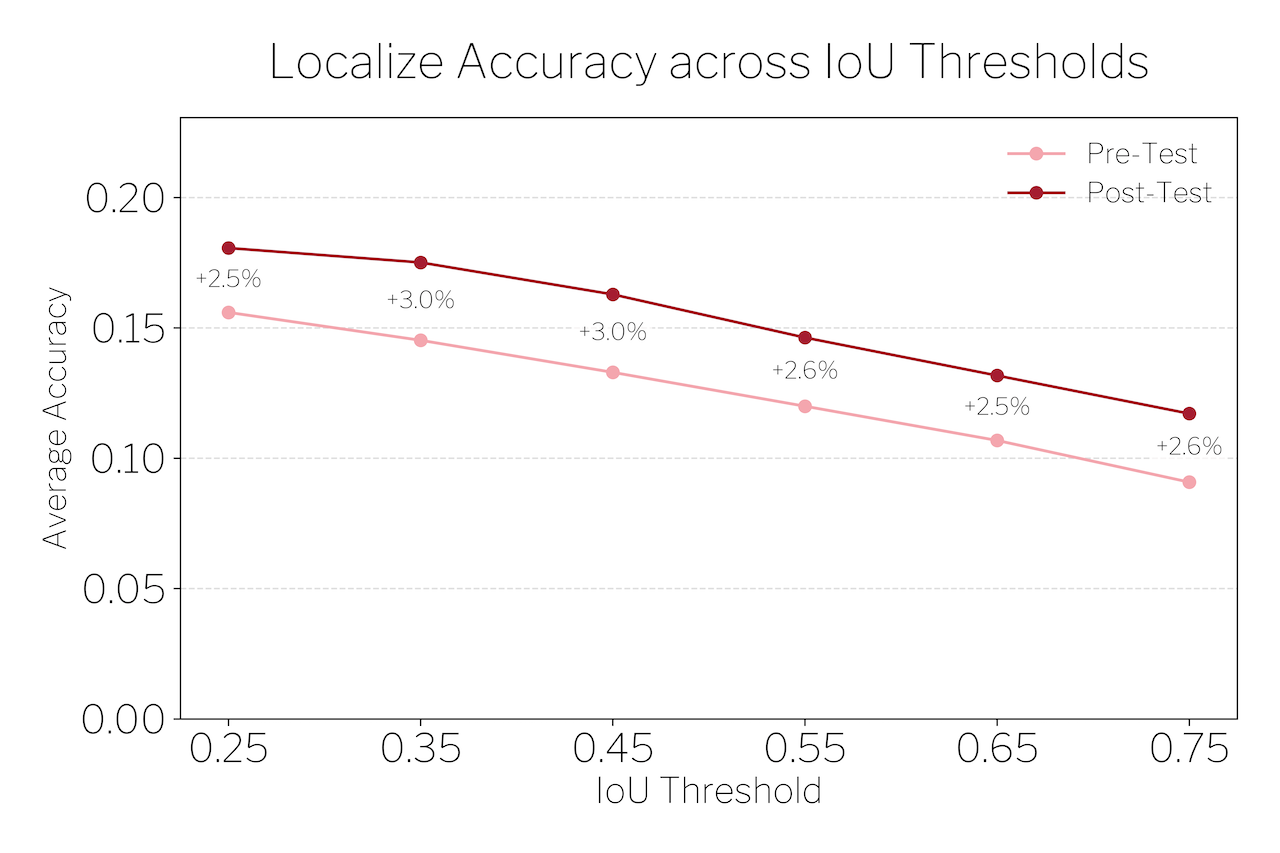}
  \captionsetup{name=Supplementary Figure}
  \caption{\textbf{Localize Accuracy Across IoU Thresholds.} Post-test accuracies remain consistently higher than pre-test across IoU thresholds.}
  \
  \label{fig:localize_thresholds}
\end{figure*}

\begin{figure*}[htbp]
  \centering
  \includegraphics[width=0.95\textwidth]{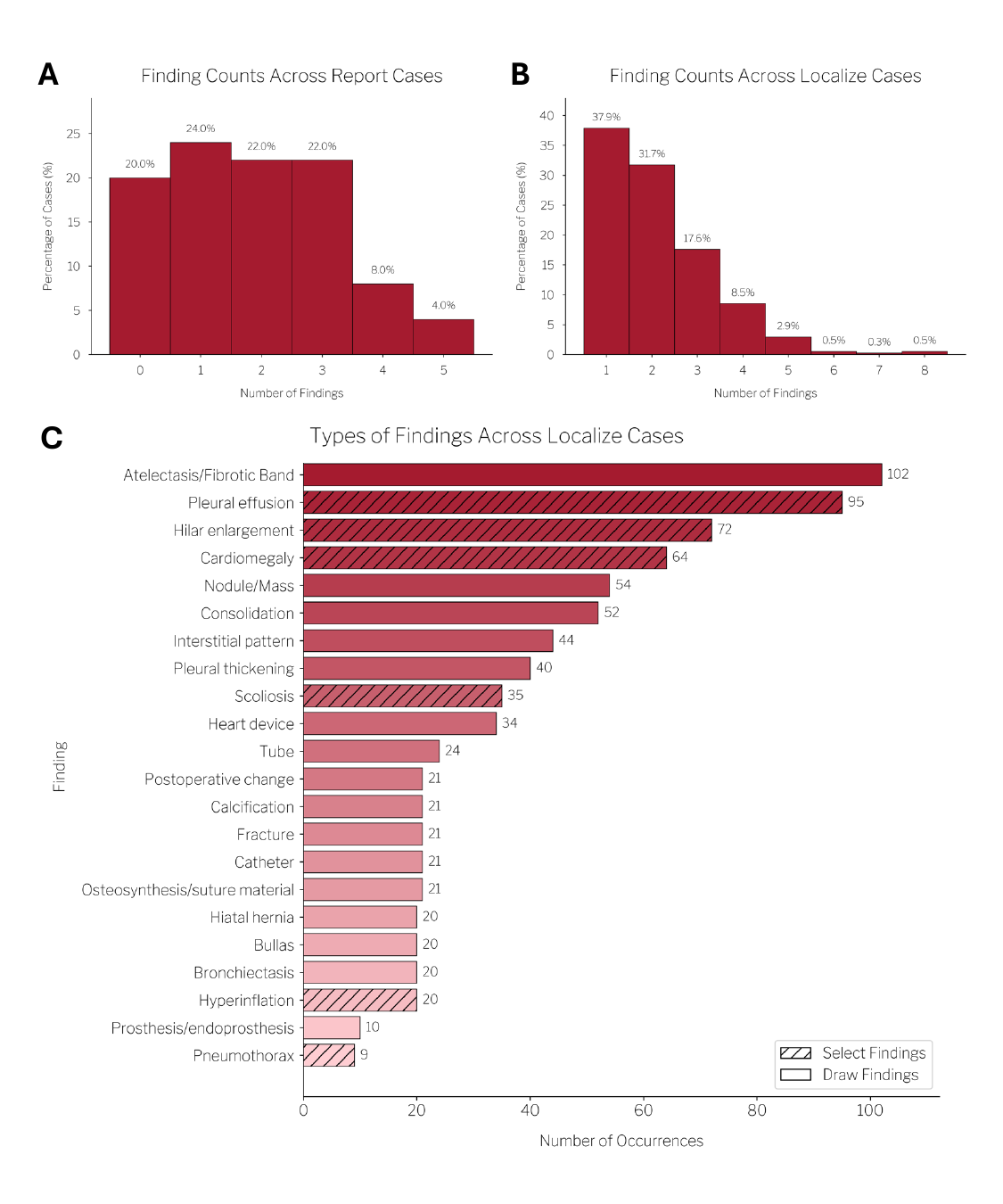}
  \captionsetup{name=Supplementary Figure}
  \caption{\textbf{Distribution of Findings and Finding Counts Across RadGame Localize and Report.} \textbf{(A)} A distribution of the number of findings present across all 150 RadGame Report cases. Cases with a diverse number of findings were selected to adjust the platform's difficulty. \textbf{(B)} A distribution of the number of findings present across all 375 RadGame Localize cases. \textbf{(C)} Counts of the different types of findings across all 375 RadGame Localize cases, divided into findings that only require a binary selection ("Select Findings") and findings that require a selection and a bounding box ("Draw Findings").}
  \
  \label{fig:distributions}
\end{figure*}

\begin{figure*}[htbp]
  \centering
  \includegraphics[width=1\textwidth]{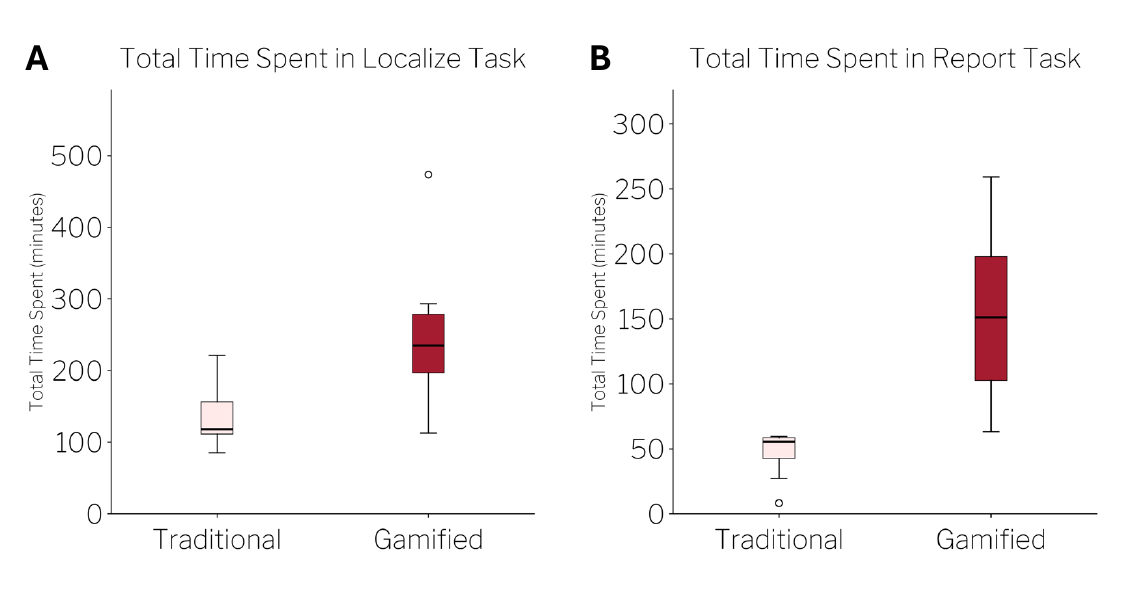}
  \captionsetup{name=Supplementary Figure}
  \caption{\textbf{Localize Accuracy Across Finding Labels.}}
  \
  \label{fig:time_distributions}
\end{figure*}

\begin{figure*}[htbp]
  \centering
  \includegraphics[width=1\textwidth]{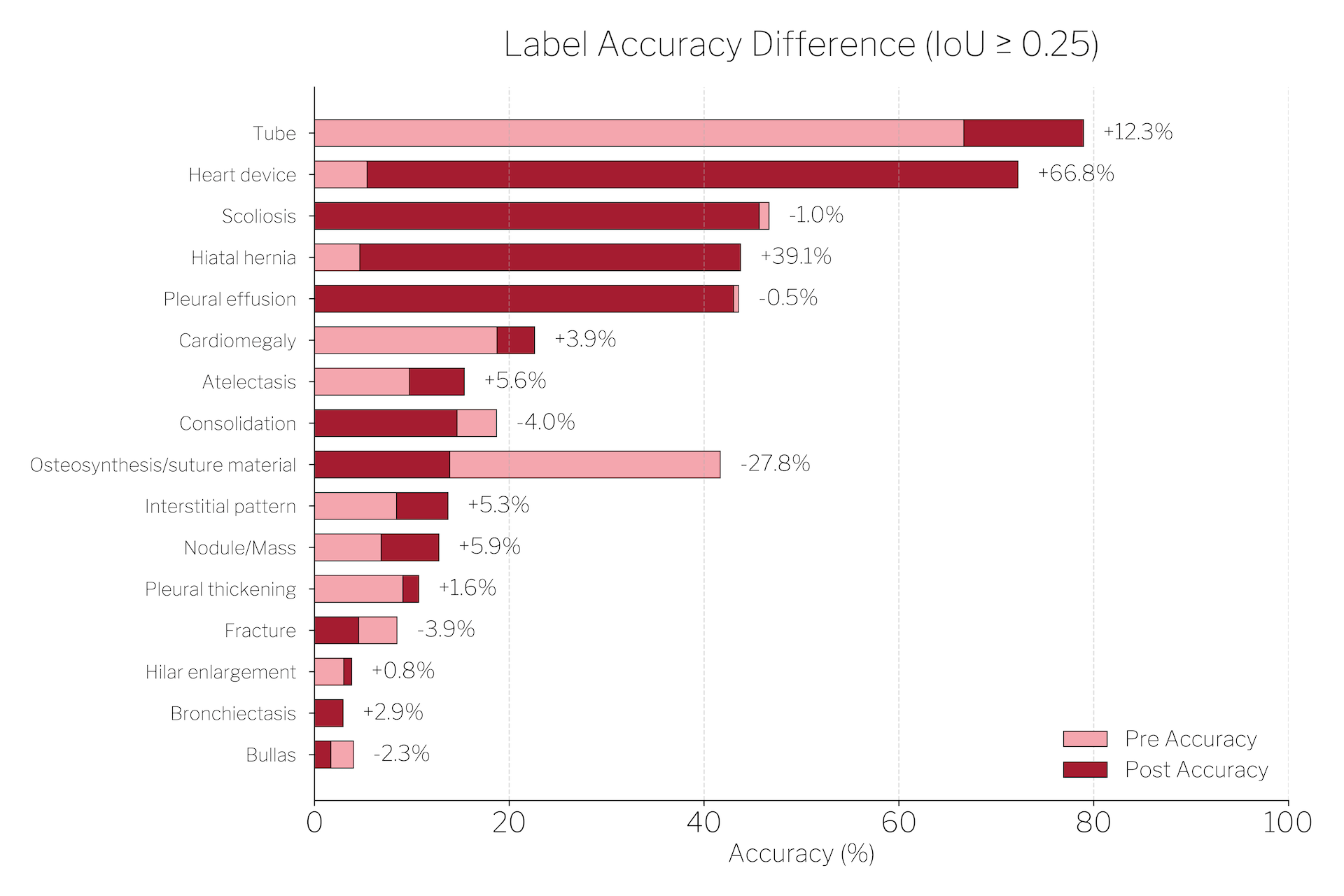}
  \captionsetup{name=Supplementary Figure}
  \caption{\textbf{Localize Accuracy Difference Pre- and Post-Test Across Finding Labels.}}
  \
  \label{fig:localize_label_accuracy}
\end{figure*}

\begin{figure*}[htbp]
  \centering
  \includegraphics[width=0.85\textwidth]{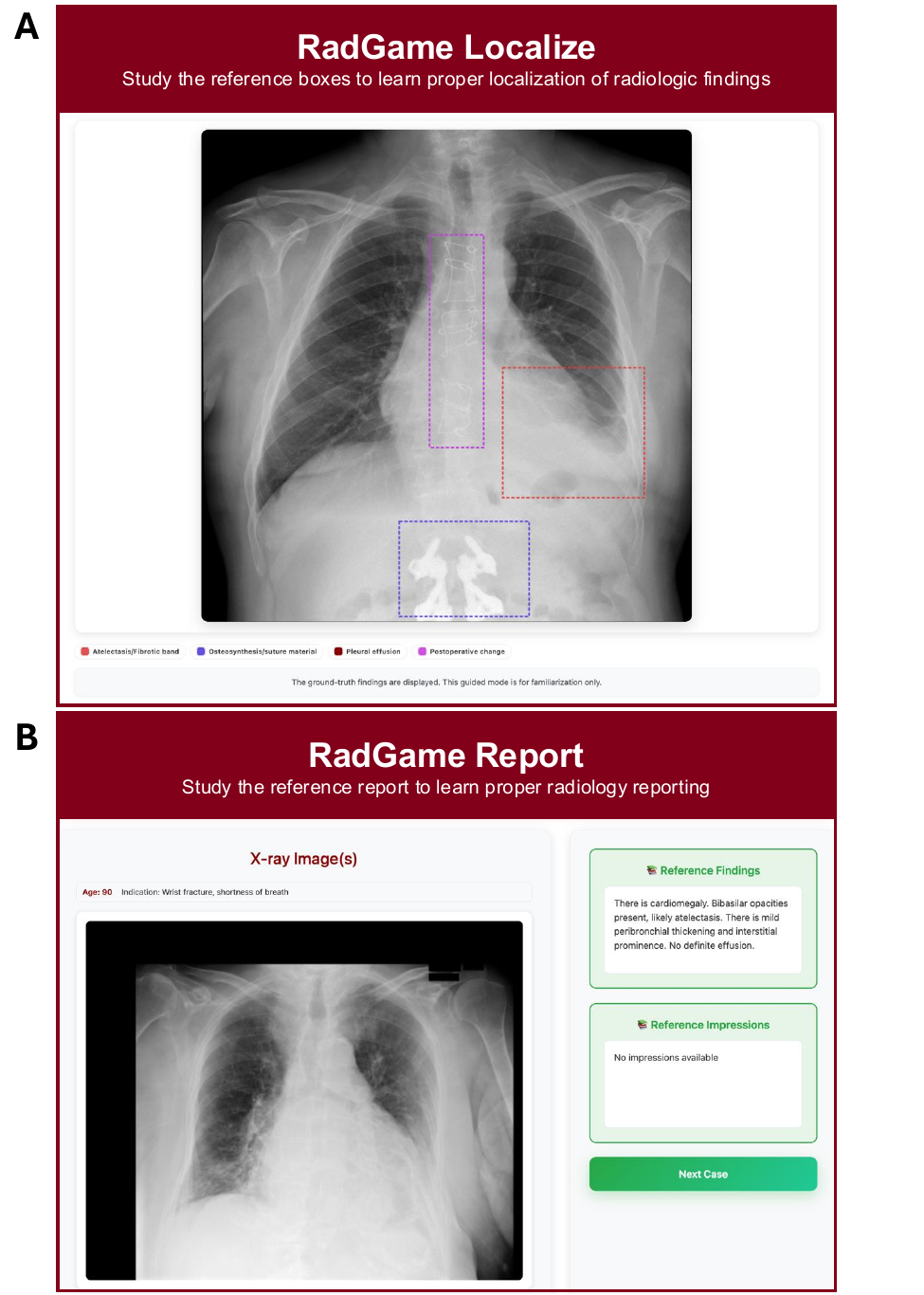}
  \captionsetup{name=Supplementary Figure}
  \caption{\textbf{RadGame User Interface (Traditional Module).} Screenshot of the RadGame platform showing modules in traditional learning mode: \textbf{(A)} \textit{Localize}, where users can see findings present in chest X-rays and their respective bounding boxes, and \textbf{(B)} \textit{Report}, where users are provided with findings from the ground truth report.}
  \
  \label{fig:screen_shots_passive}
\end{figure*}

\begin{table*}[htbp]
\centering
\captionsetup{name=Supplementary Table}
\caption{\textbf{RadGame Localize: Draw and Select Findings.}}
\label{tab:findings}
\small
\begin{tabular}{p{0.31\textwidth} p{0.31\textwidth} p{0.31\textwidth}}
\toprule
\multicolumn{3}{c}{\textbf{Draw Findings}}\\
\midrule
1. Atelectasis/Fibrotic band  & 2. Bronchiectasis           & 3. Bullas \\
4. Calcification              & 5. Catheter                 & 6. Consolidation \\
7. Fracture                   & 8. Heart device             & 9. Hiatal hernia \\
10. Interstitial pattern       & 11. Nodule/Mass             & 12. Osteosynthesis/suture material \\
13. Pleural thickening         & 14. Postoperative change    & 15. Prosthesis/endoprosthesis \\
16. Tube                      &                             &  \\
\midrule
\multicolumn{3}{c}{\textbf{Select Findings}}\\
\midrule
17. Cardiomegaly               & 18. Hilar enlargement       & 19. Hyperinflation \\
20. Pleural effusion           & 21. Pneumothorax            & 22. Scoliosis \\
\bottomrule
\end{tabular}
\end{table*}

\begin{table*}[htbp]
\centering
\captionsetup{name=Supplementary Table}
\caption{\textbf{Distribution of Cases Across Interstitial Pattern Subtypes.}}
\label{tab:interstitial_patterns}
\small
\begin{tabular}{l c}
\hline
\textbf{Interstitial Pattern Subtype} & \textbf{Number of Cases} \\
\hline
Nodular/Miliary     & 51 \\
Reticulonodular     & 133 \\
Reticular/Kerley B line   & 260 \\
\hline
\end{tabular}
\end{table*}

\begin{table*}[htbp]
\centering
\captionsetup{name=Supplementary Table}
\caption{\textbf{Comparison of Sample Accuracies Across VLMs.} We use a sample size of 100.}
\label{tab:model_accuracies}
\small
\begin{tabular}{l c c c}
\hline
\textbf{Model} & \textbf{Location Accuracy} & \textbf{Visual Accuracy} & \textbf{Class Accuracy} \\
\hline
MedGemma 4B & \textbf{88} & \textbf{96} & 90 \\
Qwen3VL 4B & 80 & 92 & 94 \\
Llama3.2 11B Vision & 81 & 86 & \textbf{96} \\
\hline
\end{tabular}
\end{table*}

\begin{figure*}[h]
\centering
\begin{tcolorbox}[
    colback=red!5,
    colframe=red!60!black,
    title=CRIMSON Prompt,
    width=\textwidth]
\footnotesize
Objective:
                Evaluate the accuracy of a candidate radiology report in comparison to a reference
                radiology report composed by expert radiologists. Only include positive findings, not normal findings. 
                Do not include notes unrelated to clinical findings.\\
                
                Process Overview:\\
                You will be presented with:
                \begin{enumerate}
                    \item The criteria for making a judgment.
                    \item The reference radiology report.
                    \item The candidate radiology report.
                    \item The desired format for your assessment.
                \end{enumerate}

                \begin{enumerate}
                    \item Criteria for Judgment:
                For each candidate report, determine only the clinically significant errors.

                Errors can fall into one of these categories:
                \begin{enumerate}[label=\alph*)]
                    \item False report of a finding in the candidate.
                    \item Missing a finding present in the reference.
                    \item Misidentification of a finding's anatomic location/position.
                    \item Misassessment of the severity of a finding.
                \end{enumerate}
                
                Note: Concentrate on the clinical findings rather than the report's writing style.
                Evaluate only the findings that appear in both reports. 
                \\ \\
                Patient Context: \\
                    Age: \{age\} \\
                    Indication: \{indication\}
                \\ \\
                IMPORTANT NOTES: 
                \begin{itemize}
                    \renewcommand\labelitemi{-} 
                    \item Evaluate only positive findings, not normal findings. If a finding is normal, it should not be counted in the errors.
                    \item Ignore all references to prior findings and studies. DO NOT COUNT THEM AS ERRORS.
                    \item Do NOT penalize the candidate report for omitting specific numeric measurements (e.g., size or dimensions of a nodule/lesion) if the underlying finding is correctly identified. Missing measurements alone is fine since the user writing the candidate report can't measure. They should only be penalized for missing the finding itself. 
                    \item Do NOT penalize omission of age-appropriate findings that are NOT clinically significant in the context of the indication and patient age. For example, if the patient is over 65 years old, do not penalize omission of expected degenerative changes such as aortic calcification, vascular tortuosity, degenerative spine changes, etc, UNLESS it is related to the indication.
                    
                    \item Do NOT hallucinate or infer findings absent from both reports.
                \end{itemize}
             
                \item  Reference Report:
                \{reference\}
           
                \item  Candidate Report:
                \{candidate\}
            
                \item  Reporting Your Assessment:
                Format your output as a JSON. Follow this specific format for your output, even if no errors are found:
                \end{enumerate}
                
                \{\\
                \hspace*{1cm}``Explanation": ``$<$Explanation$>$",\\
                \hspace*{1cm}``ClinicallySignificantErrors": \{\\
                        \hspace*{2cm}``a": [``$<$Error 1$>$", ``$<$Error 2$>$", ``...", ``$<$Error n$>$"],\\
                        \hspace*{2cm}``b": [``$<$Error 1$>$", ``$<$Error 2$>$", ``...", ``$<$Error n$>$"],\\
                        \hspace*{2cm}``c": [``$<$Error 1$>$", ``$<$Error 2$>$", ``...", ``$<$Error n$>$"],\\
                        \hspace*{2cm}``d": [``$<$Error 1$>$", ``$<$Error 2$>$", ``...", ``$<$Error n$>$"]\},\\
                    \hspace*{1cm}``MatchedFindings": [``$<$Finding 1$>$", ``$<$Finding 2$>$", ``...", ``$<$Finding n$>$"]\\
                \}
\end{tcolorbox}
\captionsetup{name=Supplementary Figure}
\caption{\textbf{The prompt used to generate CRIMSON scores.}}
\label{fig:CRIMSON-prompt}
\end{figure*}

\begin{figure*}[h]
\centering
\begin{tcolorbox}[
    colback=red!5,
    colframe=red!60!black,
    title=Style Score Prompt,
    width=\textwidth
]

    Objective: Evaluate the writing style and structure of a radiology report to determine how well it follows professional radiology reporting standards. Focus on style, structure, and systematic evaluation rather than clinical accuracy.
    
    Criteria for Judgment:\\
    
    Rate each aspect as 0 (poor), 0.5 (adequate), or 1 (excellent):

    \begin{enumerate}
        \item  SYSTEMATIC EVALUATION: Does the report cover the major chest X-ray regions?
        \begin{itemize}
            \renewcommand\labelitemi{-} 
            \item  1.0: Covers most/all major areas (lungs, heart, bones, mediastinum) in organized way
            \item 0.5: Covers several major areas but may miss 1-2 or lack organization
            \item 0.0: Only mentions 1-2 areas or very disorganized
        \end{itemize}

       \item  ORGANIZATION AND LANGUAGE: Is the report reasonably well-organized and written in appropriate clinical language?
       \begin{itemize}
        \renewcommand\labelitemi{-}
           \item 1.0: Clear organization with, complete sentences and clinical language
           \item 0.5: Some organization present, mostly complete sentences
           \item 0.0: Poor organization, incomplete sentences, non-clinical language
        \end{itemize}

    \end{enumerate}

    Candidate Report: \{candidate\}\\

    NOTES: 
        \begin{itemize}
            \renewcommand\labelitemi{-}
            \item  Do NOT recommend the user to make sections or sub-sections in the report such as Findings, Impression, etc. 
            \item  Provide 1 recommendation per scoring category that scored less than 1.0
            \item  If a category scores 1.0 (perfect), leave that recommendation field empty (``")
            \item  Keep each recommendation very concise and actionable
        \end{itemize}
    Be concise in your recommendations.
    Provide your assessment in the following JSON format:\\

    \{ \\
    \hspace*{1cm}``systematic\_evaluation\_score": $<$0, 0.5, or 1$>$, \\
    \hspace*{1cm}``organization\_language\_score": $<$0, 0.5, or 1$>$, \\
    \hspace*{1cm}``systematic\_evaluation\_recommendation": \\
    \hspace*{2cm}``$<$Recommendation if score $<$ 1, otherwise empty$>$", \\
    \hspace*{1cm}``organization\_language\_recommendation": \\
    \hspace*{2cm}``$<$Recommendation if score $<$ 1, otherwise empty$>$" \\
    \}

\end{tcolorbox}
\captionsetup{name=Supplementary Figure}
\caption{\textbf{The prompt used by Style Score to evaluate the candidate report on ``Systematic Evaluation" and ``Organization and Language".}}
\label{fig:stylescore-prompt}
\end{figure*}

\begin{figure*}[htbp]
  \centering
  \includegraphics[width=0.85\textwidth]{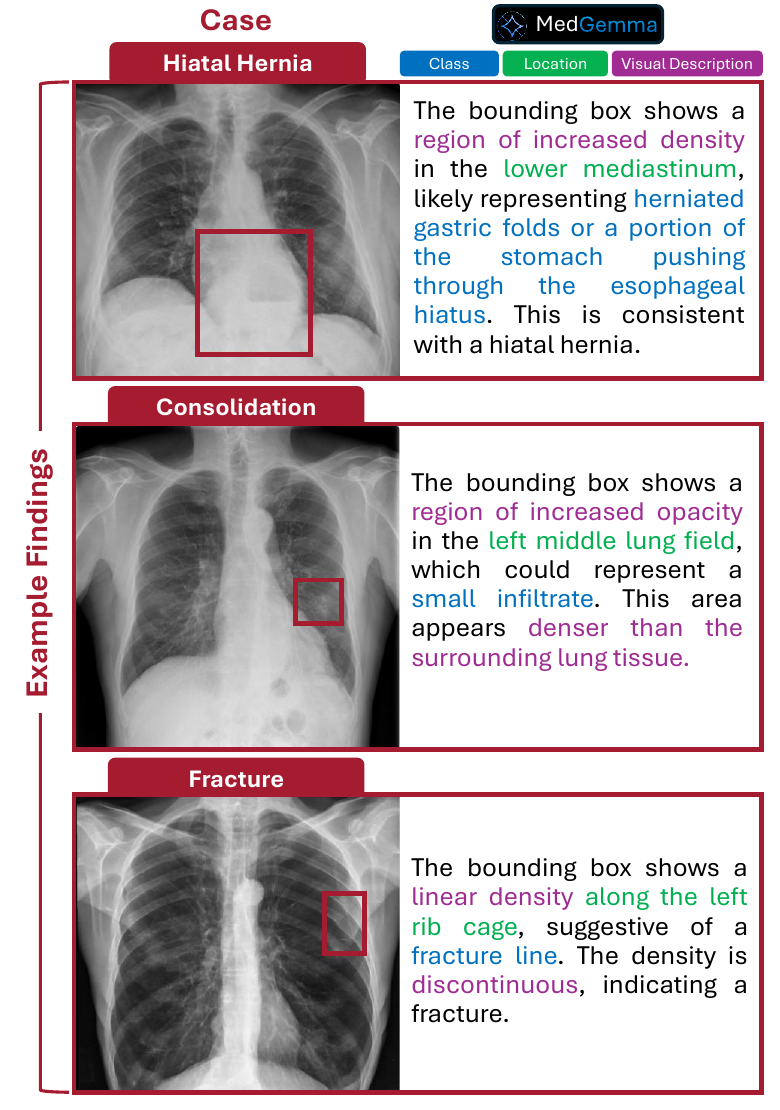}
  \captionsetup{name=Supplementary Figure}
  \caption{\textbf{Examples of MedGemma Explanations.}}
  \label{fig:med_gemma_explinations}
\end{figure*}

\begin{figure*}[htbp]
  \centering
  \includegraphics[width=0.85\textwidth]{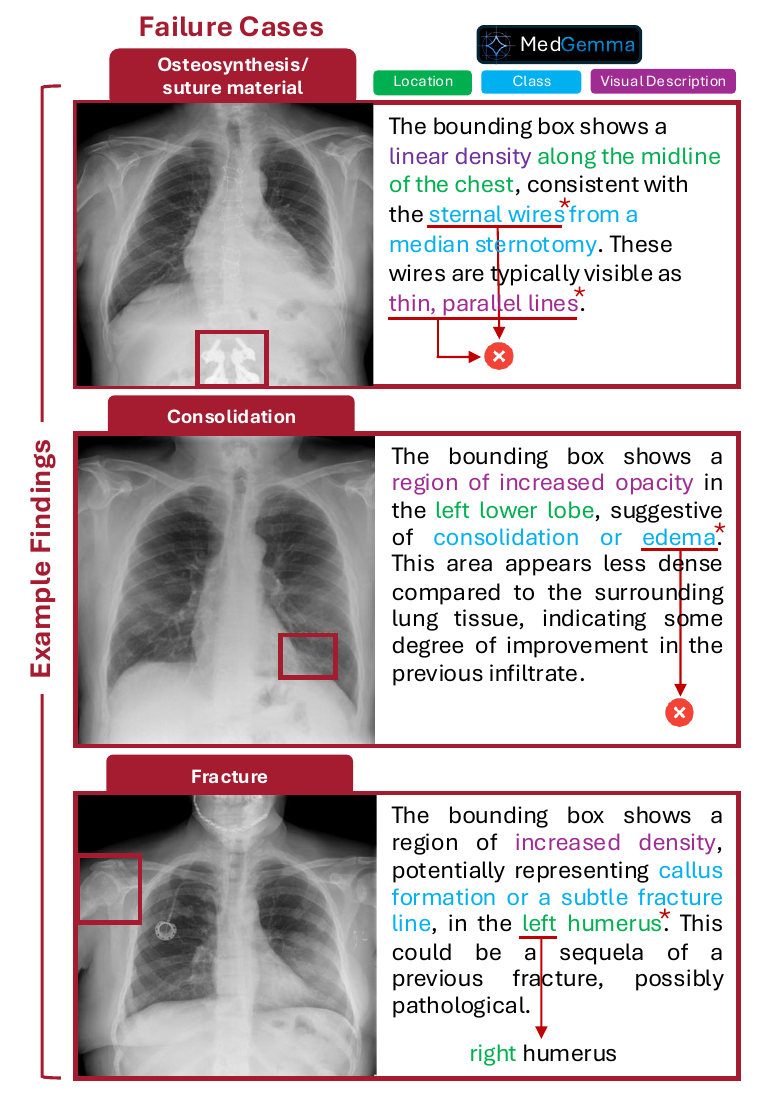}
  \captionsetup{name=Supplementary Figure}
  \caption{\textbf{Examples of MedGemma Failure Explanations.}}
  \label{fig:med_gemma_failure_explinations}
\end{figure*}

\begin{figure*}[htbp]
  \centering
  \includegraphics[width=1\textwidth]{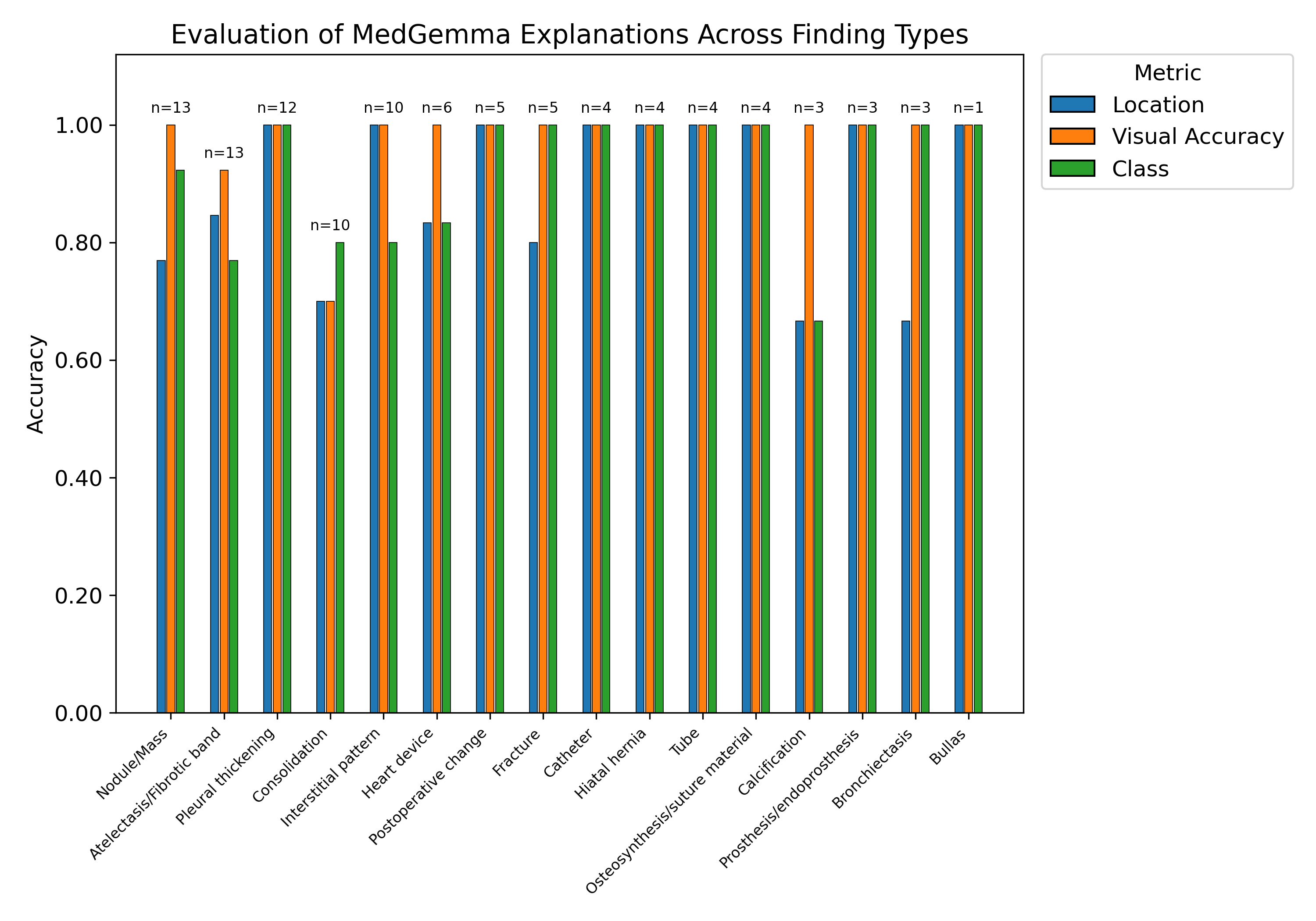}
  \captionsetup{name=Supplementary Figure}
  \caption{\textbf{Accuracy of MedGemma Explanations Across Finding Categories.} The reported accuracy is from 100 cases sampled randomly and evaluated by a senior radiologist.}
  \label{fig:explinations_accuracy}
\end{figure*}

\end{document}